%% file: WWW2020_ninghao.tex
\documentclass[sigconf]{acmart}

\usepackage{amsmath,amsfonts,amsthm,bm}

\usepackage{subcaption} 
\newcommand{\ra}[1]{\renewcommand{\arraystretch}{#1}}

\usepackage{tikz}
\usetikzlibrary{arrows,positioning,automata,calc,shapes}
\usepackage{pgfplots}
\pgfplotsset{compat=newest, scaled z ticks=false} 
\pgfplotsset{plot coordinates/math parser=false}
\newlength\figureheight 
\newlength\figurewidth

\usepackage{graphicx}
\usepackage{enumitem}
\usepackage{stmaryrd}

\AtBeginDocument{%
  \providecommand\BibTeX{{%
    \normalfont B\kern-0.5em{\scshape i\kern-0.25em b}\kern-0.8em\TeX}}}

\copyrightyear{2020}
\acmYear{2020}
\setcopyright{acmcopyright}\acmConference[CIKM '20]{Proceedings of the 29th ACM International Conference on Information and Knowledge Management}{October 19--23, 2020}{Virtual Event, Ireland}
\acmBooktitle{Proceedings of the 29th ACM International Conference on Information and Knowledge Management (CIKM '20), October 19--23, 2020, Virtual Event, Ireland}
\acmPrice{15.00}
\acmDOI{10.1145/3340531.3411919}
\acmISBN{978-1-4503-6859-9/20/10}

\begin{document}

% 1.. Explainable Recommender Systems by Design
% 2.. Towards Internally Explainable Recommender Systems
\title{Explainable Recommender Systems via Resolving \\Learning Representations}

\author{Ninghao Liu}
\affiliation{%
  \institution{Texas A\&M University, TX, USA}
%   \country{Iceland}
}
\email{nhliu43@tamu.edu}

\author{Yong Ge}
\affiliation{%
  \institution{University of Arizona, AZ, USA}
%   \country{Iceland}
}
\email{yongge@email.arizona.edu}

\author{Li Li}
\affiliation{%
  \institution{Samsung Research America, CA, USA}
%   \country{China}
}
\email{li.li1@samsung.com}

\author{Xia Hu}
\affiliation{%
  \institution{Texas A\&M University, TX, USA}
%   \country{Iceland}
}
\email{hu@cse.tamu.edu}

\author{Rui Chen}
\affiliation{%
  \institution{Samsung Research America, CA, USA}
%   \country{China}
}
\email{rui.chen1@samsung.com}

\author{Soo-Hyun Choi}
\affiliation{%
  \institution{Samsung Electronics, South Korea}
%   \postcode{78229}
}
\email{sh9.choi@samsung.com}


\begin{abstract}
Recommender systems play a fundamental role in web applications in filtering massive information and matching user interests. While many efforts have been devoted to developing more effective models in various scenarios, the exploration on the explainability of recommender systems is running behind. Explanations could help improve user experience and discover system defects. In this paper, after formally introducing the elements that are related to model explainability, we propose a novel explainable recommendation model through improving the transparency of the representation learning process. Specifically, to overcome the representation entangling problem in traditional models, we revise traditional graph convolution to discriminate information from different layers. Also, each representation vector is factorized into several segments, where each segment relates to one semantic aspect in data. Different from previous work, in our model, factor discovery and representation learning are simultaneously conducted, and we are able to handle extra attribute information and knowledge. In this way, the proposed model can learn interpretable and meaningful representations for users and items. Unlike traditional methods that need to make a trade-off between explainability and effectiveness, the performance of our proposed explainable model is not negatively affected after considering explainability. Finally, comprehensive experiments are conducted to validate the performance of our model as well as explanation faithfulness.
\end{abstract}
%given the widely available relational data recommendation scenarios


\maketitle

\section{Introduction}
\input{1_Introduction.tex}

\section{Preliminaries: Elements of Explainable Recommendation}
\input{2_Elements.tex}

\section{Problem Formulation}
\input{3_Formulation.tex}

\section{Interpretable Recommendation With Graph Data}
\input{4_Model.tex}
% Note: Be clear in what situation can we built a clustering view of data. Not in the D-dim space, but in the D/C-dim space.

\section{Experiment}
\input{5_Experiment.tex}

\section{Related Work}
\input{6_RelatedWork.tex}

\section{Conclusion and Future Work}\label{sec:append}
\input{7_Conclusion.tex}

% \begin{acks}
% The work is, in part, supported by NSF ({\#}IIS-1900990, {\#}IIS-1750074). The views and conclusions contained in this paper are those of the authors and should not be interpreted as representing any funding agencies.
% \end{acks}

%\newpage
\bibliographystyle{ACM-Reference-Format}
\bibliography{WWW2020_}

%\newpage
\appendix

\section{Appendix}
\subsection{Node Importance Estimation}
This part is to briefly introduce the computation of $s(j;u, t)$ in Section~\ref{subsec:eval_intp}. The importance computation closely follows the information propagation path in graph convolution. Specifically, if $j$ is an item node, then its importance within factor $c$ is:
\[
    s(j, c;u, t) = p(t,c)\exp(\langle \textbf{z}^{itm, c}_u , \textbf{z}^{slf}_t \rangle) \cdot \mathit{1}^{(u, j)} \frac{p(j, c)}{|\mathcal{N}_u|} ,
\]
where the first part is from the decoder, and the second half is factor affiliation degree. $\mathit{1}^{(m, j)}=1$ if $j$ connects to $u$. Here we let $s(j;u, t) = \sum_c s(j, c;u, t)$.

If $j$ is an entity node, then its importance within factor $c$ is:
\[
    s(j, c;u, t) = \exp(\langle \textbf{z}^{ent, c}_u , \textbf{z}^{ent, c}_t \rangle) \cdot \sum_{m\in |\mathcal{N}_u|}\frac{1}{|\mathcal{N}_u|} \mathit{1}^{(m, j)} \frac{p(j,c)}{|\mathcal{N}_m|} ,
\]
where the first part is from the decoder, and the second part is traced from information propagation path in GCN. Similarly, we let $s(j;u, t)$ $ = \sum_c s(j, c;u, t)$.
%\input{_Appendix}


\end{document}

%% file: 1_Introduction.tex
% RS and ERS
Recommender systems play a pivotal role in a wide range of web applications and services, in terms of distributing online content to targets who are likely to be interested in it. The majority of the efforts in the domain have been put into developing more effective models to achieve better performance. In contrast, the progress of analyzing the explainability aspect of recommender systems is running behind. Explainable recommender systems address \textit{the problem of why} -- besides providing recommendation results, they also give reasons to clarify why such results are derived~\cite{Zhang-Chen17explainableRec, yang2018towards, gunning2019darpa, Ribeiro-etal16whyshould}. Explainability may benefit a recommender system in several aspects. First, explanations help system maintainers to diagnose and refine the recommendation pipeline (system-oriented). Second, by increasing transparency, explanations promote persuasiveness and customer satisfaction of recommender systems (user-oriented).
% Our model is a generative process.

% Existing work of interpretation, limitations
Recently, Explainable AI (XAI), or interpretable machine learning, is receiving increasing attention~\cite{gunning2019darpa, Du-etal18techniques, Montavon-etal18survey} (we use "explanation" and "interpretation" interchangeably in this paper). One popular direction is post-hoc methods~\cite{Du-etal18techniques, Ribeiro-etal16whyshould}, where some examples include understanding the meaning of latent factors~\cite{peake2018explanation, Guan-etal19unifiedNLP} or reconstructing the rationale behind each prediction~\cite{yang2018towards}. However, post-hoc interpretation suffers from the issue of explanation accuracy~\cite{Rudin19pleaseStop, gunning2019darpa}. Another direction is to build explainable models or add explainable components, where attention mechanism is commonly applied to highlight important features for prediction~\cite{Seo-etal17interpretable, Wang-etal19AAAIrecomm}. Nevertheless, explanation schemes such as heatmaps in images or texts are not structured and still require subjective comprehension of users. Since effective models can learn informative latent representations from data~\cite{bengio2013representation, Higgins-etal17betaVAE, Sabour-etal17capsule}, one of the important building blocks of model explainability is the transparency of representation learning. In traditional factorization models, input features can be directly associated with latent factors to elucidate their meanings~\cite{Zhang-etal14explictFactor}, but this explanation scheme is not directly applicable to recent embedding learning frameworks.
% Meanwhile, convoluted structures of advanced models still impede us from further understanding the internal model mechanics.
%  as well as the inability of further improving models
% As a result, it poses extra difficulty for users to be persuaded by explanations, or for developers to quickly identify the drawbacks of the system. 

In this work, we propose an explainable recommendation model through promoting the transparency of latent representations. To begin with, we summarize three elements that help make a model more interpretable. We name them as IOM elements since they involve certain requirements on the \textbf{I}nput data format, \textbf{O}utput attribution, and \textbf{M}iddle-level representations. The proposed explainable model is designed based on the IOM elements. Users, items and attribute entities are processed as nodes in a graph. Furthermore, the efforts on interpretable recommendation are split into three parts: (1) we disentangle the \textit{interactions} between latent representations in different layers; (2) multiple semantic \textit{factors} are identified automatically from data; (3) latent dimensions are divided into \textit{segments} according to their information source (i.e., node types) and affiliated factors. The first part is achieved via graph convolutional networks (GCNs). However, different from previous work~\cite{Hamilton-etal18survey, Wang-etal19KGAT, Wang-etal19KGCN}, we propose to understand GCNs from another perspective by comparing its working mechanism with those of fully-connected networks~\cite{Cheng-etal16wideanddeep} and capsule networks~\cite{Sabour-etal17capsule}. The second and third aspects are achieved through a novel architecture design, where different dimensions of latent representations focus on different aspects of data. Different from previous work~\cite{Epasto-etal19isaSingle, Liu-etal19isaSingle, Wang-etal19multipleEmbd}, factor discovery and representation learning are jointly conducted. In this way, we are able to depict how information flows from input features through these latent states to prediction results. Different from some existing work that sacrifices effectiveness for interpretability, the proposed model still achieves good performance in the experimental evaluation. Finally, besides visualizing explanations, we also quantitatively measure explanation accuracy.

Our contributions in this work are summarized as follows:
\begin{itemize}[leftmargin=*]
    \item We propose an explainable recommendation model that improves transparency of latent representations. Specifically, we propose to unravel the interactions between representations, and segment the latent dimensions into different factors. The consideration of interpretability does not negatively influence model effectiveness. 
    \item We summarize and refine the elements that improve the interpretability of recommendation. These elements involve certain requirements on the input data format, output attribution, and middle-level representations, and are thus named as IOM elements. The proposed model is designed based on IOM elements.
    \item We conduct comprehensive experiments to evaluate the proposed model in two aspects, effectiveness and interpretability. We show that the effectiveness of the model will not be hindered by interpretability. Also, we quantitatively measure the accuracy of interpretation through adversarial attack. 
    %We provide codes at: \url{https://github.com/ninghaohello/IntpRecSysRRep}.
\end{itemize}

%Recent advances in deep learning have motivated various models that leverage reviews for building recommender systems~\cite{Tay-etal18multiPointer, Seo-etal17interpretable, Zheng-etal17jointReviews, Cheng-etal18aspectLatentFactor}. Limitations: (1) It is problematic to concatenate all reviews to represent users and items; (2) Unable to incorporate other side information; (3) Interpretation are highlighted on words, and the information is not aggregated to a compact form, so still requires another step of human comprehension and summary. 

%Not only interpretable, but also allow straightforward human intervention into the operation of recommender systems. Existing models that directly manipulate on texts may be able to provide explanation through word-level or phrase-level heatmaps. However, it is not straightforward to adjust the information flow if defects are observed by human experts.

%% file: 2_Elements.tex
In this section, we discuss the important elements in a recommendation pipeline that could promote human understanding. The proposed model, which is to be formally introduced in the next section, considers all of these three elements. It is worth noting that some directions of XAI may not focus on these elements, and they are beyond the discussion of this paper.
% These elements involve a number of requirements on input data, attributable output, and middle-level representations. 

\subsection{Conceptualization of Input}
\label{sec:concept} 
Besides complexity and parameter sizes, deep models are commonly regarded as black-boxes because they operate on low-level features, rather than high-level, discrete, human-comprehensible concepts. For examples, computer vision models operate on pixels, and natural language models take word vectors as input. The direct consequence is that there is a lack of concrete carriers to concisely depict information flows from input to output. Some recent research starts to explore how to conceptualize data into discrete interpretation basis, and derive explanations or explainable models upon it~\cite{ZhangQ-etal18cnnTrees, Yao-etal18gcnImgCaption, Kim-etal18concepts, Zhou-etal18visualBasis}. 

In this paper, we also work on discretized input data. In recommender systems, users and items can naturally be regarded as discrete concepts. For side content information, we make use of knowledge graphs (KGs) where each entity in a KG can be regarded as a concept. Developing new techniques to extract concepts from raw input, or building more informative graphs, is not a main contribution of this paper. Finally, we connect users and items with observed interactions, items and entities if they are related, entities and entities if they are already connected in a KG. We ignore the link types in a KG. Therefore, users, items and KG entities are built into a graph used as input into recommender systems.

%Different from traditional feature engineering + feature embedding work in that several features may together influence the same embedding segments.

%(1) Knowledge graph only describe the objective aspects of items or concepts which may not be used as the clue for inferring the subjective preferences of users. (2) In certain application scenarios such as movie recommendation, it is beneficial to apply knowledge graphs, since the main elements (e.g., actors, directors, genres) are readily well-known so that they could be found in existing knowledge graphs. For some other scenarios such as restaurant recommendation or hotel recommendation, the information could be found in structured data.

%in~\cite{Zhang-Chen17explainableRec} may be seen another instantiation of the three elements above, where relevant users, features, text and visual patterns can be regarded as attributing prediction to input of different modalities, while word cluster construction can be regarded as building concepts with words.

\begin{figure}[t]
\centering
\includegraphics[scale=0.41]{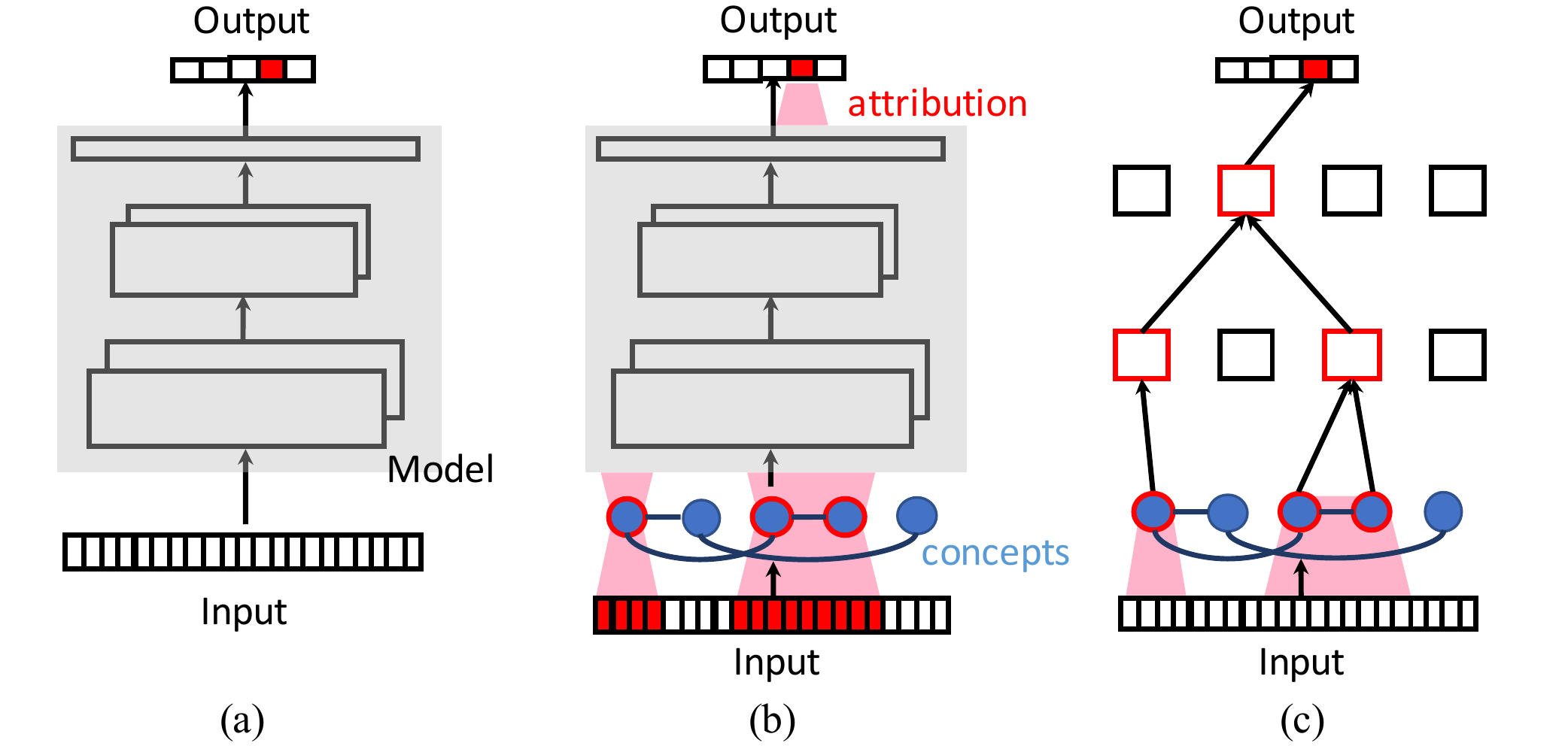}
\vspace{-15pt}
\caption{Models with different degrees of transparency. (a): A black-box model. (b): A model with conceptualized input and attributable output. (c): A model that extends (b) with transparent representation learning.}
\label{fig:elements}
\vspace{-8pt}
\end{figure}

\subsection{Attributable Output} 
A basic requirement of model interpretability is the ability to obtain the attribution of features to a  prediction output (i.e., output is attributable to certain features). Depending on whether the attribution is obtained after or along with the prediction, interpretation techniques can be categorized into post-hoc methods~\cite{Simonyan-etal13deepInsideCNNsaliency, Smilkov-etal18smoothgrad, Ribeiro-etal16whyshould, yang2018towards} and building intrinsically explainable models~\cite{Velickovic-etal18graphAttention}. Post-hoc methods usually approximate the original model with simpler but explainable ones, or resort to backpropagation to estimate the model's sensitivity to different features. Post-hoc methods have the advantage of being model-agnostic~\cite{Ribeiro-etal16whyshould}, but they have also been criticised as not being faithful enough in reproducing the working mechanism of the original model~\cite{Rudin19pleaseStop}.
In this work, our goal is to design an \textit{explainable model}, instead of developing new post-hoc methods, to provide human comprehensible rationales for recommendation.
%In contrast, attention methods usually attach additional components to the original model, such as attention weights, to trace the contribution of features. 

\subsection{Resolved Middle-Level Representations} 
The resultant model, after considering the two elements above, can be roughly depicted in Figure~\ref{fig:elements}(b). It is better than the fully black-box model in Figure~\ref{fig:elements}(a), but still does not shed light on the model's internal mechanism.

\begin{figure}[t]
\centering
\includegraphics[scale=0.56]{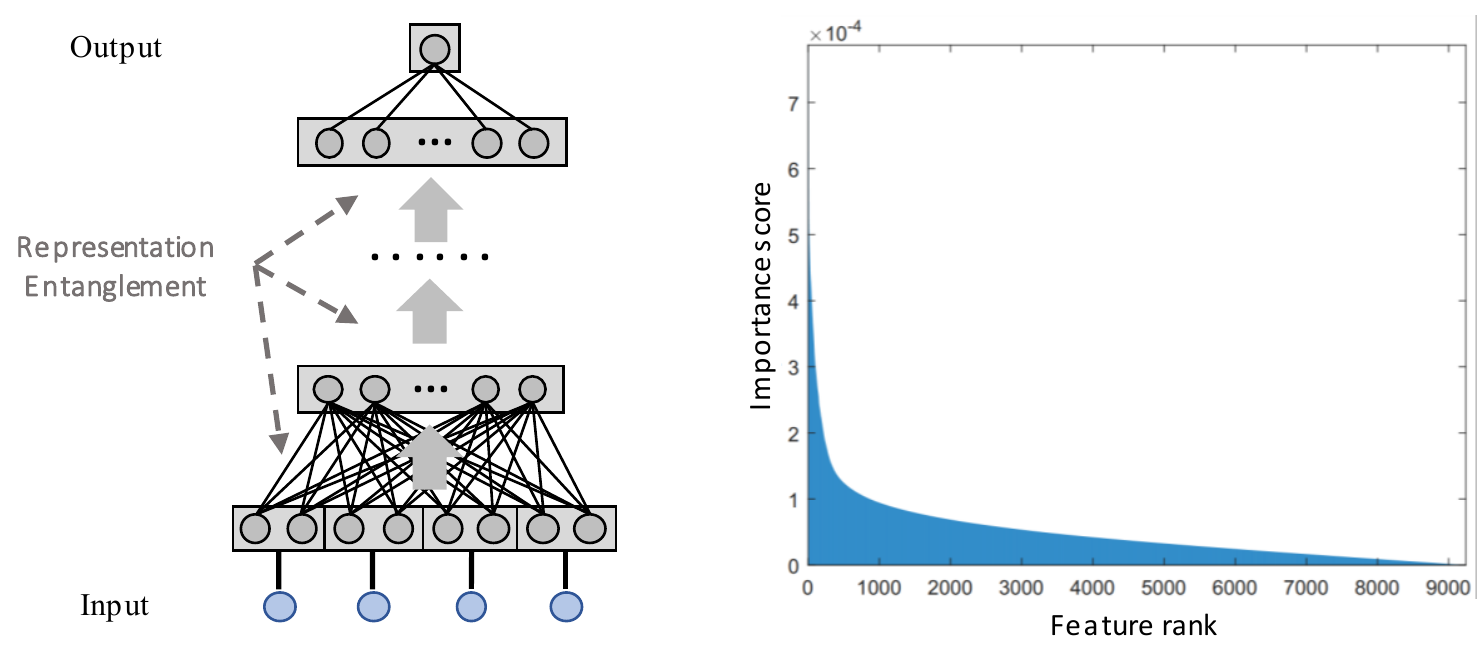}
\vspace{-3pt}
\caption{Feature interaction entangling in MLP.}
\label{fig:mlp}
\vspace{-10pt}
\end{figure}

The main difficulty for better interpreting complex models lies in disentangling and understanding the interactions between latent representations. An example is shown in Figure~\ref{fig:mlp} where we build a recommendation model based on a Multi-Layer Perceptron (MLP) with three hidden layers with $L_1$ regularization on the LastFM dataset~\cite{Wang-etal19KGCN}. The feature importance explanation of each prediction is computed using the denoised gradient-based method~\cite{Smilkov-etal18smoothgrad}. We then normalize the importance scores to the unit sum, and sort the features according to their scores. We randomly select $500$ samples from test data and compute the average of the sorted scores, as shown in the right part of Figure~\ref{fig:mlp}. It is observed that, although a small number of features are important for each prediction, the effect (area under the curve) of the long tail of less important features together is still significant. The reason for this phenomenon is that a fully-connected network tends to entangle features without explicitly separating the interactions of latent dimensions~\cite{Tsang-etalNIPS18interactionTransparency}.
%To tackle the issues above, some models constrain feature interactions in certain schemes, such as Factorization Machines~\cite{rendle2010factorization} (FM) and its variants~\cite{Xiao-etal17attentionFM}, but they may be limited in expressiveness.
Without tackling this challenge, the efforts of conceptualizing input and attributing output would be largely thwarted, since contributions of different features will be indistinguishable.
%Without tackling this challenge, the efforts of conceptualizing input and attributing output would be largely thwarted. First, even if we can guarantee that each feature has a concrete meaning, entanglement of representations will make contributions of different features indistinguishable. Second, although it is not hard to obtain attentions or importance scores of features as post-hoc explanations, convoluted representations will make the explanations hard to be understood by end users or utilized by system developers.

\subsubsection{Disentanglement of Representation Interactions} From the analysis above, we are facing the dilemma where we hope to both: (1) explore \textit{flexible} feature interactions, and (2) constrain feature interactions to be \textit{concise} to trace. These requirements remind us of the recent Capsule Network (CapsNet)~\cite{Sabour-etal17capsule}. Each "capsule" represents a certain concept, and a dynamic routing process is run to decide the contribution between capsules of adjacent layers. Let $\textbf{z}_{j}$ and $\textbf{z}_{i}$ denote the embeddings of a higher-level capsule $v_j$ and a lower-layer capsule $v_i$. The capsules interact in a way that
\begin{equation}\label{eq:capsule}
    \textbf{z}_{j} = \text{squash}(\sum_{i=1}^I a_{i,j} \textbf{z}_{j|i}), \,\,\,\,\,\,\, \textbf{z}_{j|i} = \textbf{W}_{i,j} \textbf{z}_i,
\end{equation}
where $\textbf{W}_{i,j}$ is the bilinear mapping matrix, "squash" is a non-linear function, and $a_{i,j} \ge 0$ is the contribution score tracking the relation between capsules. Some recent work~\cite{Li-etal19CARP, Zhao-etal18capsuleText, Sabour-etal17capsule, Zhang-Zhu18survey} claims to use CapsNet to improve model interpretability. Nevertheless, several challenges impede us from directly applying CapsNet to certain recommendation scenarios. First, if we treat users or items as capsules, then the number of parameters $\{\textbf{W}_{i,j}\}$ would be too large. Second, dynamic routing significantly increases time complexity and memory. Third, how to instill structured knowledge from human experts into the model remains unsolved.
% in terms of identifying semantic concepts (i.e., $\textbf{z}_{i}$, $\textbf{z}_{j}$) and exploring information flows (i.e., $a_{i,j}$)

Next we will show that graph convolution processes information in a similar manner to CapsNet, and the challenges faced by CapsNet can be easily tackled by GCN. Let $\textbf{z}_j$ be the embedding of node $v_j$. Through graph convolution, its embedding is updated as:
\begin{equation}\label{eq:gcn}
    \textbf{z}_j \leftarrow \sigma(\sum_{v_i \in \mathcal{N}_j\cup\{v_j\}} a_{i,j} \textbf{W}\textbf{z}_i) ,
\end{equation}
where $\mathcal{N}_j$ denotes the neighbors of $v_j$, $\sigma(\cdot)$ is the activation function, $a_{i,j}$ is the attention score. By comparing Eq(\ref{eq:capsule}) and Eq(\ref{eq:gcn}), it can be observed that both models use vectors to represent objects, and apply the weighted vector sum to propagate information. In this sense, GCN may be seen as a special case of CapsNet, where iterative routing is omitted and bilinear mapping reduces to a single matrix $\textbf{W}$ that reduces the number of parameters. Moreover, GCN can naturally handle relational data and human knowledge organized in the form of graphs.
%In graphs, we do not differentiate node IDs when matching them with a mapping matrix, so using one matrix $\textbf{W}$ in each layer is a natural choice and it reduces the parameter size. 
%To sum up, GCNs maintain interpretability by propagating concise feature information flows towards prediction. The information contained in a node representation can be traced back to low-level features through attention. 
In this paper, we use graph convolution as the basic operation to build explainable recommender systems.

\subsubsection{Segmentation of Latent Dimensions}
Another aspect of representation learning transparency is to explicitly separate the information captured by different dimensions of representations. There are two advantages for latent dimension separation. First, each set of dimensions corresponds to one semantic aspect, so we are able to learn meaningful representations. Second, disentanglement of latent dimensions may improve model effectiveness via encoding more nuanced relations in data~\cite{Epasto-etal19isaSingle}. For example, if we use an embedding to depict the interest of a user, then ideally we would want different dimensions to separately capture user interests in different item genres, such as clothes, electronics, and daily necessities.
% Besides being viewed as dividing each embedding into several segments, the proposed model can also be seen as learning multiple embeddings for each node.

Disentangling latent dimensions is a nontrivial task despite some preliminary work proposed recently. \cite{Epasto-etal19isaSingle, Liu-etal19isaSingle} extend network embedding by learning several vectors for each node. Wang \textit{et al.}~\cite{Wang-etal19multipleEmbd} utilize item categories as part of the supervised information to associate latent dimensions with different aspects. However, how to deal with extra knowledge or automatically discover factors without category information are beyond their discussion. 
%In the following section, we will introduce the details of the proposed model in achieving dimension separation in graph convolution.

% In MLP, the high-level neurons usually creates high-order interactions that may not make much sense from human perspective~\cite{Tsang-etalNIPS18interactionTransparency}.

% The interpretation aspect of GNNs can also be related with that of Capsule Networks, which recently receive a lot of attention partly due to the claim of improved model interpretability~\cite{Sabour-etal17capsule, Zhao-etal18capsuleText}. In capsule networks, each capsule, represented by a vector, is associated with certain visual or semantic meaning. Through learning, connections between adjacent low-level capsules and high-level capsules are established, thus propagating information from input to output. Similarly, in GNNs, each node can be seen as a capsule, and graph convolution aggregate information from low layers to high layers. More technical details for relating the two models can be found in later sections.

%% file: 3_Formulation.tex
%\textcolor{red}{should we at least mention knowledge graph somewhere in Section 2, 3 and 4? Now the concept of knowledge graph first appears in Section 5. It seems a bit abrupt.}

\textbf{Notations.} In this paper, column vectors are represented by boldface lowercase letters, such as $\mathbf{a}$ and $\mathbf{b}$; matrices are represented by boldface capital letters, such as $\mathbf{W}$ and $\mathbf{Y}$; sets are represented by calligraphic capital letters, such as $\mathcal{U}$ and $\mathcal{N}$. The $i$-th entry of a vector $\mathbf{b}$ is denoted by $b_i$, and entry $(i, j)$ of a matrix $\mathbf{W}$ is denoted by $W_{i,j}$. 
%The $i$-th row and $j$-th column are denoted by $W_{i,\cdot}$ and $W_{\cdot,j}$, respectively.
As discussed above, the input into recommender systems is constructed as a graph $\mathcal{G}$ with nodes denoted by $\mathcal{V}$. We include three types of nodes in the graph, i.e., users, items and knowledge entities, where $\mathcal{U}$, $\mathcal{T}$ and $\mathcal{E}$ denote the user set, item set and entity set, respectively. Here note that $v\in \mathcal{V} = \mathcal{U} \cup \mathcal{T} \cup \mathcal{E}$ can be used to denote any type of nodes when the node type is not specified. In this paper, we consider three types of links in our scenario, user-item links $\{(u, t)\}$, item-entity links $\{(t, e)\}$ and entity-entity links $\{(e, e')\}$, where $u\in \mathcal{U}$, $t\in \mathcal{T}$ and $e, e'\in \mathcal{E}$. A user-item link $(u, t)$ indicates an observed interaction (e.g., purchase, impression) between user $u$ and item $t$. An item-entity link $(t, e)$ unveils the item's quality or attribute described by entity $e$. An entity-entity link $(e, e')$ simply stores the relation between the two entities. Without loss of generality, it is also possible to add user-entity links $\{(u, e)\}$ to reflect a user's attention or attitude towards an entity, but such information is not directly available in our experimental data. There is also a ground-truth interaction matrix $\mathbf{Y} \in \mathbb{R}^{|\mathcal{U}|\times|\mathcal{T}|}$, where $Y_{u,t}=1$ means that there is an observed interaction between user $u$ and item $t$, and $Y_{u,t}=0$ otherwise.
%A user-concept link $(u, e)$ reflects the user's attention or attitude towards concept $e$.
% We do not consider link types in this work.

\noindent \textbf{Problem Definition.} Given the input graph $\mathcal{G}$, our goal is to predict whether user $u$ is interested in an item $t$. That is, we aim to learn a prediction function $\hat{Y}_{u,t} = f(u, t|\mathcal{G}, \Theta)$, where $\hat{Y}_{u,t}$ is the estimated probability that user $u$ is interested in item $t$, and $\Theta$ denotes the parameters to learn. More importantly, additional constraints for explainability are added to the architecture of $f$.

%% file: 4_Model.tex
In this section, we introduce the proposed explainable recommender system, which is based on the IOM elements introduced in previous sections. First, we extend the traditional explainable factor decomposition model into a more general representation learning framework. Second, we introduce the recommendation interpretation schema. Third, we introduce a novel graph convolution architecture to resolve latent dimensions, as well as the variational autoencoder based learning framework.

\begin{figure}[t]
\centering
\includegraphics[scale=0.55]{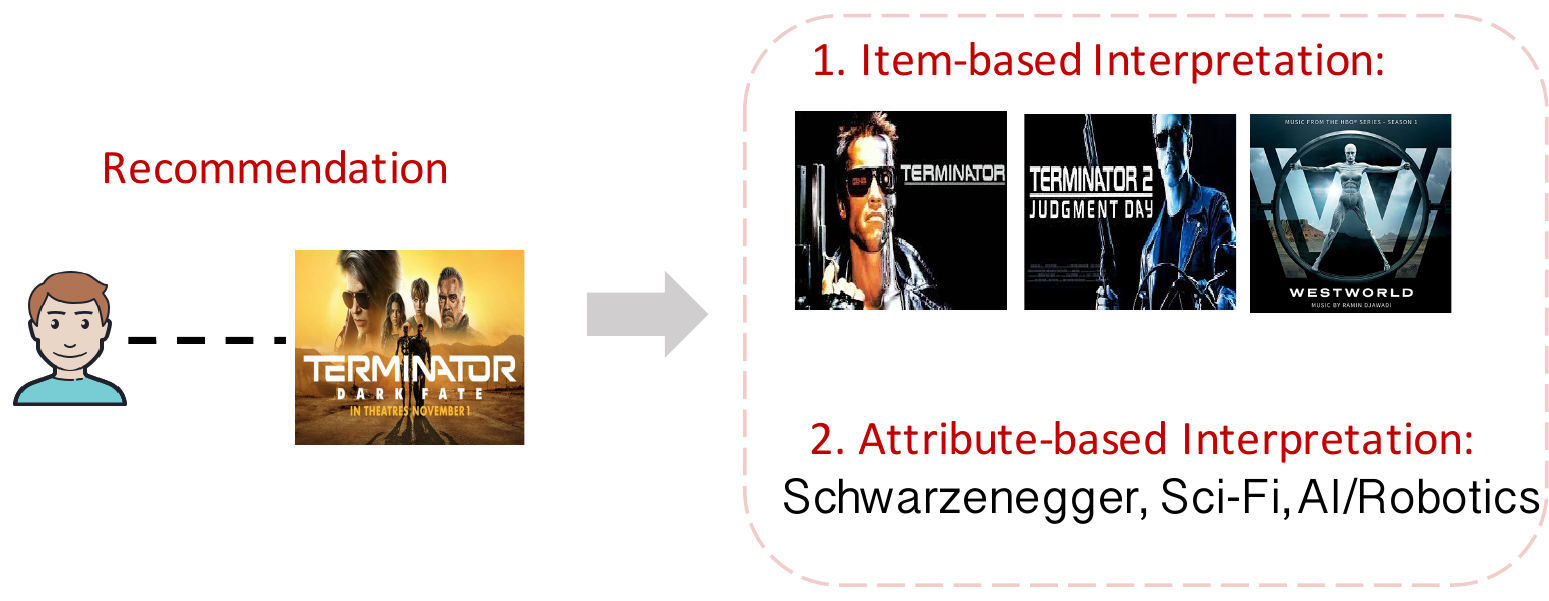}
\vspace{-10pt}
\caption{Interpretation schema. A user's interest is explained through the historical items he has interacted with, as well as knowledge entities.}
\label{fig:input}
\vspace{-5pt}
\end{figure}

\begin{figure*}[t]
\centering
\includegraphics[scale=0.48]{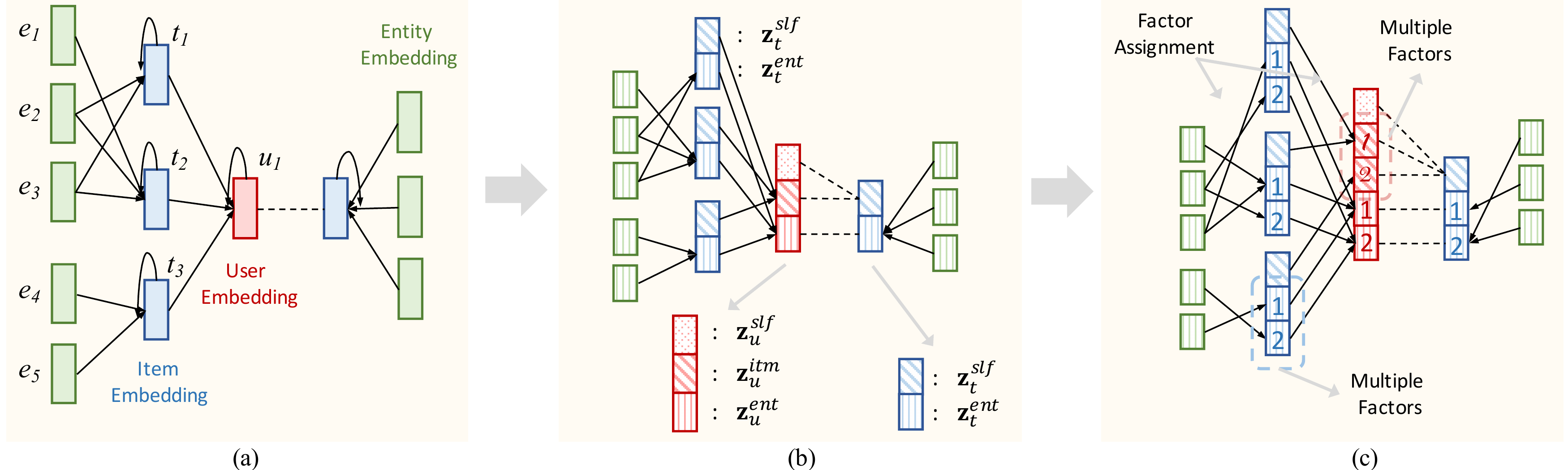}
\vspace{-5pt}
\caption{Overview of the proposed model design procedure, best viewed with color. (a): Traditional GCN information propagation. (b): GCN after discriminating information sources. (c): GCN after resolving information sources and factors.}
\label{fig:main}
\vspace{0pt}
\end{figure*}
% In (c), each user owns two sets of factors, respectively inherited from entities and items. Thus, we use two types of font faces for factor IDs to distinguish the two sources.

\subsection{Node-Factor Factorization}
In each recommendation application, we assume that there exist a number of latent factors which compactly describe different aspects of the nodes in the graph. Suppose there are $C$ factors. In traditional factorization models~\cite{Koren-etal09MF}, we consider the parameterization where a nonnegative score $F_{v,c}$ is assigned for each pair of node $v\in \mathcal{V}$ and factor $c\in \{1, ..., C\}$ to measure their affiliation strength. More specifically, for a user node $u\in \mathcal{U}$, $F_{u,c}$ represents the user's attention or interest on factor $c$. For an item node $t\in \mathcal{T}$, $F_{t,c}$ quantifies the degree that factor $c$ can be used to describe item $t$. Therefore, the tendency that factor $c$ independently triggers the interaction between $u$ and $t$ could be computed as $F_{u,c} \cdot F_{t,c}$. Then, an observed link $Y_{u, t}$ can be decomposed as $Y_{u, t} \approx \sum^C_{c=1} F_{u,c} \cdot F_{t,c}$. The explanation behind is that user $u$ is more likely to interact with item $t$ if the item has high quality or reputation on more factors which are also highly valued by the user.

Traditionally, $F_{u,c}$ and $F_{t,c}$ are directly solved as nonnegative parameters optimized through matrix factorization~\cite{kong2011robust, Koren-etal09MF, Zhang-etal14explictFactor}. However, the design of explicit factors is not fully compatible with recent representation learning frameworks. To combine the explainability of explicit factorization and the flexibility of recent learning frameworks, we embed different factors within nodes separately into the latent space. That is, we use a vector $\textbf{z}^c_v\in \mathbb{R}^D$ to represent node $v$ described by factor $c$, where $D$ is the embedding dimension of a single factor. The score between user $u$ and item $t$ under factor $c$ is thus computed as $\langle \textbf{z}^c_u, \textbf{z}^c_t \rangle$, where $\langle\cdot\,,\cdot\rangle$ denotes inner product. And the link factorization is thus formulated as $Y_{u, t} \approx \sum^C_{c=1} \langle \textbf{z}^c_u, \textbf{z}^c_t \rangle $. The link prediction can also be written as $Y_{u, t} \approx \langle \textbf{z}_u, \textbf{z}_t \rangle $, where $\textbf{z}_v = [\textbf{z}^1_v;\textbf{z}^2_v;...;\textbf{z}^C_v]\in \mathbb{R}^{C\times D}$.
%For example, in movie recommendation, each factor could be associated with certain movie categories or features, so we are able to discover users' interests.
%These factors are compact elements for describing user and item groups, and can be explicitly designed to associate with input features for explanation purposes~\cite{Zhang-etal14explictFactor}. 
For each observed user-item pair, it is expected that only a part of the factors will be activated in preserving similarity. If factor $c$ is the potential cause of making user $u$ interact with item $t$, in order for $\langle \textbf{z}^c_u, \textbf{z}^c_t \rangle $ to have a large score, not only should the angle be small between ${\textbf{z}^c_u}$ and $\textbf{z}^c_t$, we also expect them to have large lengths. 
%Therefore, the length of an embedding segment, i.e., $||\textbf{z}^c_v||_2$, reflect the affiliation degree between node $v$ and factor $c$.

In certain models, more flexible interactions between factorized representations are designed. For example, in Factorization Machines~\cite{rendle2010factorization}, cross-factor interactions are allowed (i.e., $\langle {\textbf{z}^{c_1}_u}, \textbf{z}^{c_2}_t \rangle$ also contributes to $Y_{u,t}$ and $c_1\neq c_2$). In addition, if we assume that users have multiple interests but items are constrained with a single property, then $Y_{u, t} \approx \sum^C_{c=1} \langle \textbf{z}^c_u, \textbf{z}_t \rangle$.

%Besides learning representations for graph nodes, we also propose to embed the $C$ factors into latent space, so that each factor $c$ is represented by an embedding $\mathbf{z}_c$. Our model is designed in the way that, for each node $v$, $\mathbf{z}^c_v$ is related with $\mathbf{z}_c$. Also, $\mathbf{z}_c$ is jointly learned with the embeddings of nodes under factor $c$. The details of model construction are discussed in the below sections.

\subsection{Interpretation Hierarchy}
In this work, we treat user nodes, item nodes and entity nodes as \textit{concepts}. Different concepts are assigned to different levels. A concept on level $\ell$ is represented by concepts on level $\ell'$ where $\ell' < \ell$. Specifically, let $v$ and $v'$ denote concepts on level $\ell$ and $\ell'$, then we could interpret the embedding of $v$ as
$
    \textbf{z}_{v} \approx \sum_{v'} p_{v, v'}\textbf{z}_{v'} ,
$
where $p_{v, v'}$ denotes the contribution from $v'$ to $v$. Low-level concepts tend to be simple, while high-level concepts are complex and composite. Lower-level concepts can be regarded as the \textit{basis} to represent higher-level ones. Note that besides $\ell' = \ell-1$, any $\ell' < \ell$ can be chosen as the basis level to represent concepts on level $\ell$.

The levels of concepts are defined in an iterative manner. The user nodes are on the top level $\ell^{max}$. The item nodes are on level $\ell^{max}-1$ as they connect to users. The entities which connect to items are on level $\ell^{max}-2$. In general, if a node $v$ connects to another node on level $\ell$ and $v$ does not belong to a level higher than $\ell$, then it belongs to level $\ell-1$. In the end, the bottom entity nodes are on level $0$. In this work, to simplify illustration, we assume that there is only one level of entity nodes, so that $\ell^{max}=2$ for user nodes. An example of recommendation explanation for a user is given in Figure~\ref{fig:input}. The proposed model can be easily extended to deal with multi-level knowledge entities.

\subsection{Resolving Information Flow} 
We now introduce how to segment $\mathbf{z}_v$ to different factors. Under traditional GCNs shown in Figure~\ref{fig:main}(a) and Eq(\ref{eq:gcn}), the embeddings of user $u$ and item $t$ on the final GCN layer are computed as:
\begin{equation}
\begin{split}
    \textbf{z}_u &\leftarrow \texttt{COMBINE}(\textbf{z}_u, \texttt{AGGREGATE}(\{\textbf{z}_t: t \in \mathcal{N}_u\subseteq \mathcal{T}\})) , \\
    \textbf{z}_t &\leftarrow \texttt{COMBINE}(\textbf{z}_t, \texttt{AGGREGATE}(\{\textbf{z}_e: e \in \mathcal{N}_t\subseteq \mathcal{E}\})) .
\end{split}
\end{equation}
It is worth noting that we restrict the information flow direction. An item embedding $\textbf{z}_t$ receives information from knowledge entities, as well as its own embedding on the lower GCN layer. Similarly, a user embedding $\textbf{z}_u$ receives information from itself on the lower layer, the items that have been interacted, and knowledge entities as second order connections. Finally, top-layer $\textbf{z}_u$ and $\textbf{z}_t$ engage in computing recommendation prediction $\hat{Y}_{u,t}$. Common $\texttt{COMBINE}()$ operations include summation or concatenation followed by transformation, while common $\texttt{AGGREGATE}()$ includes summation and mean operations~\cite{Ying-etal18GCNrecomm, Hamilton-etal18graphSAGE, Wang-etal19KGAT}. In the proposed model, we modify traditional \texttt{COMBINE}() and \texttt{AGGREGATE}() operations to resolve information propagation between representations. 

\subsubsection{Modification to COMBINE()} 
Instead of merging information from lower-level representations, we keep $\textbf{z}_v$ and $\texttt{AGGREGATE}(\{\textbf{z}_{v'}: v' \in \mathcal{N}_v\})$ separate if $v$ and $v'$ are different types of nodes. Specifically, as shown in Figure~\ref{fig:main}(b), we define item embedding as:
\begin{equation}
    \begin{split}
        \textbf{z}_t &= \textbf{z}^{slf}_t \mathbin\Vert \textbf{z}^{ent}_t, \\
        \text{where}\,\,\,\,\,\, \textbf{z}^{ent}_t &= \texttt{AGGREGATE}(\{\textbf{z}^{slf}_e: e \in \mathcal{N}_t\subseteq \mathcal{E}\}).
    \end{split}
\end{equation}
Here $\textbf{z}^{slf}_t$ denotes the item itself's bottom-level embedding, and $\mathbin\Vert$ means concatenation. 
Similarly, for each user embedding $\textbf{z}_u$, it receives information from both items and knowledge entities that describe those items, so we define it as:
\begin{equation}
    \begin{split}
        \textbf{z}_u &= \textbf{z}^{slf}_u \mathbin\Vert \textbf{z}^{itm}_u \mathbin\Vert \textbf{z}^{ent}_u , \\
        \text{where}\,\,\,\,\,\, \textbf{z}^{itm}_u &= \texttt{AGGREGATE}(\{\textbf{z}^{slf}_t: t \in \mathcal{N}_u\subseteq \mathcal{T}\}), \\
        \textbf{z}^{ent}_u &= \texttt{AGGREGATE}(\{\textbf{z}^{ent}_t: t \in \mathcal{N}_u\subseteq \mathcal{T}\}) .
    \end{split}
\end{equation}
Here $\textbf{z}^{itm}_u$ describes the user through the historical items he has interacted with. $\textbf{z}^{ent}_u$ describes the user with knowledge entities.

\subsubsection{Modification to AGGREGATE()} Furthermore, instead of directly aggregating information from items or entities without distinguishing their natures or semantics, we first assign low-level embeddings into different factors, and then send information accordingly. Specifically, as shown in Figure~\ref{fig:main}(c), we define item-side aggregation as:
\begin{equation}\label{eq:zent}
    \begin{split}
        \textbf{z}^{ent}_t = \bigparallel^{C_1}_{c=1}\, \textbf{z}^{ent, c}_t,\,\,\,\,\,\, \text{where}\,\,\, \textbf{z}^{ent, c}_t &= g\big(\sum_{e \in \mathcal{N}_t} \frac{p(e, c)}{|\mathcal{N}_t|}\, \textbf{z}^{slf}_e  \big)         \,.
    \end{split}
\end{equation}
Here $p(e, c)\in [0,1]$ denotes the affiliation degree between entity $e$ and factor $c$. Also, we define user-side aggregation as:
\begin{align}\label{eq:zuser}
        \textbf{z}^{itm}_u &= \bigparallel^{C_2}_{c=1}\, \textbf{z}^{itm, c}_u,\,\,\,\,\,\, \text{where}\,\,\, \textbf{z}^{itm, c}_u = g\big(\sum_{t \in \mathcal{N}_u} \frac{p(t, c)}{|\mathcal{N}_u|}\, \textbf{z}^{slf}_t \big)         \,, \\
        \textbf{z}^{ent}_u &= \bigparallel^{C_1}_{c=1}\, \textbf{z}^{ent, c}_u,\,\,\,\,\,\, \text{where}\,\,\, \textbf{z}^{ent, c}_u = \sum_{t \in \mathcal{N}_u} \frac{1}{|\mathcal{N}_u|} \textbf{z}^{ent, c}_t         \,.
\end{align}
Here $p(t, c)\in [0,1]$ denotes the affiliation degree between item $t$ and factor $c$. $g$ is a nonlinear mapping module. The design above is based on the principle that: (1) $\textbf{z}^{slf}_v$ will first be assigned to some factors $c$ according to $p(v,c)$ scores, and then contribute to the corresponding higher-level embeddings after the nonlinear mapping; (2) For those embeddings that are already factorized (i.e., $\textbf{z}^{ent, c}_t$), they will directly contribute to higher-level counterparts (i.e., $\textbf{z}^{ent, c}_u$) within the same factor $c$. It is worth noting that we include two index sets $\{c|1\leq c\leq C_1\}$ and $\{c|1\leq c\leq C_2\}$ because they index different factors.
%Also, $g_1$ and $g(2)$ are important for the effectiveness of the proposed model.
Also, if we have more layers of entity nodes, their embeddings could be learned just via normal GCNs without applying our method, because entity nodes in different layers are still of the same type.

Till now, the learnable parameters $\Theta$ in our model include: $\textbf{z}^{slf}_e$, $\textbf{z}^{slf}_t$, $\textbf{z}^{slf}_u$, $p(e,c)$ and $p(t,c')$ for $e\in \mathcal{E}$, $t\in \mathcal{T}$, $u\in \mathcal{U}$, $1\leq c\leq C_1$ and $1\leq c'\leq C_2$.

\subsection{The Proposed Model Architecture}
We now introduce the model details for learning node embeddings, estimating factor affiliation scores, and making predictions.
%Here $\mathbf{z}_c$ actually acts as the \textit{anchor vector} for factor $c$. Also, $similarity(\mathbf{z}^c_v, \mathbf{z}_c | \bm{\sigma}_c)$ is a rescaling scalar, where a large value indicates that factor $c$ is representative in describing the properties of node $v$.
In general, the recommendation model is based on a variational autoencoder (using other encoder-decoder schemes may also work, but is beyond our discussion). The model details are as below.
%Finally, we introduce how to obtain explanation for each user-item prediction.

\subsubsection{Encoder.} The encoder returns a user embedding $\textbf{z}_u$ and an item embedding $\textbf{z}_t$ for each interaction. Encoder modules are based on graph convolution designed in the above subsection, with the new \texttt{COMBINE()} and \texttt{AGGREGATE()} operations. To estimate factor affiliation degrees $p(e, c)$ and $p(t, c)$, we maintain two sets of embeddings $\{\textbf{z}^{ent}_c | 1 \leq c\leq C_1 \}$ and $\{\textbf{z}^{itm}_c | 1 \leq c\leq C_2 \}$ as learnable parameters for each factor $c$. Then we define
\begin{align}
    p(e, c) &= softmax({\cos (\textbf{z}^{slf}_e, \textbf{z}^{ent}_c)}/\gamma) ,\,\,\, \\
    p(t, c) &= softmax({\cos (\textbf{z}^{slf}_t, \textbf{z}^{itm}_c)}/\gamma), 
\end{align}
where $softmax()$ normalizes over different factors, $\gamma$ is a hyper-parameter to skew distributions as a form of self-training~\cite{Xie-etal16unsupervised}. For example, making $\gamma<1$ amplifies the gap between strong and weak assignments. Through explicitly parameterizing factor embeddings, we reduce the number of parameters than treating $p(e, c)$ or $p(t, c)$ as independent coefficients over different argument pairs.
% \begin{align}
%     p(e, c) = \frac{{\cos (\textbf{z}^{slf}_e, \textbf{z}^{ent}_c)}/\gamma}{\sum^{C_1}_{c'=1} {\cos (\textbf{z}^{slf}_e, \textbf{z}^{ent}_{c'})}/\gamma} ,\,\,\,
%     p(t, c) = \frac{{\cos (\textbf{z}^{slf}_t, \textbf{z}^{itm}_c)}/\gamma}{\sum^{C_2}_{c'=1} {\cos (\textbf{z}^{slf}_t, \textbf{z}^{itm}_{c'})}/\gamma}, 
% \end{align}

During the training process, we adopt the variational autoencoder paradigm~\cite{Kingma-Welling15autoVarBayes}. The generation of $\textbf{z}_u$ and $\textbf{z}_t$ are not deterministic, but are under certain probabilistic distribution. Also, we let different segments be mutually independent, so distributions are defined as below:
\begin{align}
    q(\textbf{z}_t | \mathcal{G}) &= q(\textbf{z}^{slf}_t | \mathcal{G})\cdot \prod^{C_1}_{c=1} q(\textbf{z}^{ent,c}_t | \mathcal{G}), \\
    q(\textbf{z}_u | \mathcal{G}) &= q(\textbf{z}^{slf}_u | \mathcal{G}) \cdot \prod^{C_2}_{c=1} q(\textbf{z}^{itm,c}_u | \mathcal{G}) \cdot \prod^{C_1}_{c=1} q(\textbf{z}^{ent,c}_u | \mathcal{G}) .
\end{align}
The probability of each segment follows a multivariate normal distribution. According to the designed information flow, $q(\textbf{z}^{ent,c}_t | \mathcal{G}) = N(\bm{\mu}^{ent,c}_t, \text{diag}(({\bm{\sigma}^{ent,c}_t})^2))$, where by following Eq(\ref{eq:zent}), we define $\bm{\mu}^{ent,c}_t = g^{\mu}(\sum_{e \in \mathcal{N}_t} \frac{p(e, c)}{|\mathcal{N}_t|}\, \textbf{z}^{slf}_e)$, and $\bm{\sigma}^{ent,c}_t = g^{\sigma}(\sum_{e \in \mathcal{N}_t} \frac{p(e, c)}{|\mathcal{N}_t|}\, \textbf{z}^{slf}_e)$. The generation of other segments are similar, so we skip them here.

\subsubsection{Decoder.} The decoder computes the similarity between $\textbf{z}_u$ and $\textbf{z}_t$ jointly over various factors, as illustrated in Figure~\ref{fig:main}(c):
\begin{equation}\label{eq:yut}
\begin{split}
    p(Y_{u, t} | \textbf{z}_u, \textbf{z}_t) &\propto \exp(\langle \textbf{z}^{slf}_u , \textbf{z}^{slf}_t \rangle) + \sum^{C_1}_{c=1}p(t,c) \exp(\langle \textbf{z}^{itm, c}_u , \textbf{z}^{slf}_t \rangle) \\
    &+ \sum^{C_2}_{c=1} \exp(\langle \textbf{z}^{ent, c}_u , \textbf{z}^{ent, c}_t \rangle) ,
\end{split}
\end{equation}
where $\langle\cdot\,,\cdot\rangle$ denotes inner product. The first term can be seen as the traditional collaborative filtering component. The second term contributes to prediction based on user historical interactions, where the embedding space contains $C_1$ factors. The third term contributes to prediction based on knowledge entities, where the embedding space contains $C_2$ factors. In this work, however, we discard the first term because it hurts the interpretability of the model. Also, it is worth noting that $p(t, c)$ in the second term is necessary, since it scales $\textbf{z}^{slf}_t$ corresponding to different factors.

\subsubsection{Learning.} The overall likelihood is decomposed into the sum of likelihoods of individual interactions as $\log p(\textbf{Y}) = \sum_{u,t} p(Y_{u,t})$. In practice, we optimize its variational lower bound $L_{\text{ELBO}}$~\cite{Kingma-Welling15autoVarBayes}:
\begin{equation}
\begin{split}
    L_{\text{ELBO}}(Y_{u,t}) =\, &\mathbb{E}_{q(\textbf{z}_u, \textbf{z}_t | \mathcal{G})} \log p(Y_{u,t} | \textbf{z}_u, \textbf{z}_t) \\
    &- \text{KL}[q(\textbf{z}_u, \textbf{z}_t | \mathcal{G}) || p(\textbf{z}_u, \textbf{z}_t)],
\end{split}
\end{equation}
where $\text{KL}[\cdot||\cdot]$ is the Kullback-Leibler divergence. The definition of $p(Y_{u, t} | \textbf{z}_u, \textbf{z}_t)$ could be found in Eq(\ref{eq:yut}) with normalization over all item candidates. We also assume that $\textbf{z}_u$ and $\textbf{z}_t$ are generated independently, so $p(\textbf{z}_u, \textbf{z}_t) = p(\textbf{z}_u)\cdot p(\textbf{z}_t)$ and $q(\textbf{z}_u, \textbf{z}_t | \mathcal{G}) = q(\textbf{z}_u | \mathcal{G}) \cdot q(\textbf{z}_t | \mathcal{G})$. We further take Gaussian prior, where $p(\textbf{z}_u) = N(\textbf{z}_u | \textbf{0}, \textbf{I})$ and $p(\textbf{z}_t) = N(\textbf{z}_t | \textbf{0}, \textbf{I})$. $q(\textbf{z}_u | \mathcal{G})$ and $q(\textbf{z}_t | \mathcal{G})$ are generated by encoders as introduced before.

%% file: 5_Experiment.tex
In this section, we evaluate the proposed model on several real-world datasets. First, we evaluate the effectiveness of the proposed model. Second, we quantitatively measure the accuracy of interpretation, and analyze the insight from interpretation.

\subsection{Datasets and Metrics}
\begin{itemize}[leftmargin=*]
    \item \textbf{MovieLens}\footnote{https://grouplens.org/datasets/movielens/}: A benchmark dataset from movie recommendation. Original ratings are converted into binary scores, indicating whether a user has interacted with a movie~\cite{He-etal17NCF}.
    \item \textbf{LastFM}\footnote{https://grouplens.org/datasets/hetrec-2011/}: A dataset obtained from the Last.fm online music website. Only interactions between users and music artists are used. The social network information is not included.
    \item \textbf{Yelp}\footnote{https://www.yelp.com/dataset/challenge/}: A dataset adopted from Yelp challenge. Restaurants and bars are used as items. Attributes in the dataset are used as knowledge entities. Each user and item have at least $10$ interactions.
\end{itemize}
Each dataset is associated with a knowledge graph as side information released in~\cite{Wang-etal19KGCNLS, Wang-etal19KGAT}. The datasets' statistics are available in Table~\ref{table:datasets}. For each dataset, we randomly hold out $200$ users for validation and $200$ users for testing. For both groups of users, $80\%$ of interactions are randomly sampled and put into training data, while the remaining are put into validation and testing data. The model to be tested has the best performance on the validation dataset.

The hyperparameter settings of the proposed model on each dataset are listed in Table~\ref{table:datasets}. Specifically, $D$ denotes the dimension of each factor's embedding. $C$ denotes the total number of factors, so that the total embedding dimension is $2C\times D$ for each user node, $(C+1)\times D$ for each item node, and $D$ for each entity node. $lr$ is the learning rate. $l_2$ weight is the regularization weight for model parameters. Batch size corresponds to the number of users put into models for training, so that the number of batches in each epoch is $\#$users$/$batch\_size. For the decoder part in Eq(\ref{eq:yut}), instead of using $\textbf{z}^{slf}_t$ and $\textbf{z}^{ent, c}_t$ from GCN, a more effective solution is to maintain another embedding dictionary as learnable parameters to replace them~\cite{ma2019learning}. Such a replacement is beneficial for model performance. The interpretation for user interest is not affected. The reported experimental results are based on this practical modification.
The effectiveness of models is evaluated under two metrics, $Recall@k$ ($k=2, 10, 50, 100$) and $NDCG@100$. 
% Specifically, the $Recall@K$ is defined as:
% \begin{equation}
%     Recall@K = \frac{|\{\text{top K retrieved items}\}\cap\{\text{ground-truth items}\}|}{|\{\text{ground-truth items}\}|}.
% \end{equation}
% And the $NDCG@K$ is defined as:
% \begin{equation}
%     NDCG@K = Z_K \sum_{i=1}^K \frac{2^{r_i}-1}{log_2(i+1)},
% \end{equation}
% where $Z_K$ is for normalization to ensure that the perfect ranking has the value of $1$. $r_i=1$ if the $i$-th retrieved item is within ground-truth, and otherwise $r_i=0$.

\begin{table}[t]
	\centering
      \begin{tabular}[0.1\textwidth]{c|ccc}
      \toprule
      \multicolumn{1}{c}{\bfseries Dataset} & \multicolumn{1}{c}{MovieLens} & \multicolumn{1}{c}{LastFM} & \multicolumn{1}{c}{Yelp}\\
      \hline 
      $\#$users & $138$,$159$  & $1$,$872$ & $45$,$919$  \\
      $\#$items & $16$,$954$  & $3$,$846$ & $45$,$538$ \\ 
      $\#$entities & $85$,$615$  & $5$,$520$ & $1$,$341$ \\ 
      $\#$edges & $13$,$501$,$622$ & $42$,$346$ &  $1$,$185$,$068$ \\
      \hline 
      $D$ & $30$  & $16$ & $25$ \\
      $C$ & $4$ & $4$ & $6$ \\
      $\gamma$ & $0.1$ & $0.1$ & $0.1$ \\
      $lr$ & $4\times10^{-4}$  &  $2\times10^{-4}$ & $2\times10^{-4}$  \\
      $l_2$ weight & $10^{-8}$ & $10^{-8}$  & $5\times10^{-9}$  \\
      batch\_size & $512$ & $128$  & $256$  \\
      epoch & $10$ & $100$  & $15$  \\
      \bottomrule
      \end{tabular}
     \vspace{-0pt}
	\caption{Statistics of the datasets.} \label{table:datasets}
\vspace{-5pt}
\end{table}

\begin{table*}[t]
\normalsize
\centering
\setlength{\tabcolsep}{2.9pt}
\ra{1.}
\begin{tabular}{@{}c|cccc|cccc|cccc@{}}\toprule
\multicolumn{1}{c}{} &
\multicolumn{4}{c}{\textbf{MovieLens}} & \multicolumn{4}{c}{\textbf{LastFM}} & \multicolumn{4}{c}{\textbf{Yelp}} \\
\cmidrule(l){2-5} \cmidrule(l){6-9} \cmidrule(l){10-13}
\multicolumn{1}{c}{} & 
\multicolumn{1}{c}{\textbf{R@2\,\,}} & \multicolumn{1}{c}{\textbf{R@10\,}} & \multicolumn{1}{c}{\textbf{R@50\,}} & \multicolumn{1}{c}{\textbf{R@100}} & \multicolumn{1}{c}{\textbf{R@2\,\,}} & \multicolumn{1}{c}{\textbf{R@10\,}} & \multicolumn{1}{c}{\textbf{R@50\,}} & \multicolumn{1}{c}{\textbf{R@100}} & \multicolumn{1}{c}{\textbf{R@2\,\,}} & \multicolumn{1}{c}{\textbf{R@10\,}} & \multicolumn{1}{c}{\textbf{R@50\,}} & \multicolumn{1}{c}{\textbf{R@100}} \\ \midrule
{\bf NMF} & $0.064$ & $0.219$ & $0.435$ & $0.556$   & $0.118$ & $0.247$ & $0.428$ & $0.520$   & $0.021$ & $0.051$ & $0.113$ & $0.152$    \\
{\bf FM} & $0.074$ & $0.224$ & $0.432$ & $0.551$   & $0.120$ & $0.250$ & $0.436$ & $0.532$   & $0.022$ & $0.053$ & $0.169$ & $0.228$    \\
{\bf CKE} & $0.070$ & $0.221$ & $0.445$ & $0.571$   & $0.121$ & $0.243$ & $0.443$ & $0.548$   & $0.020$ & $0.060$ & $0.177$ & $0.260$  \\
{\bf RippleNet} & $0.185$ & $0.242$ & $0.479$ & $0.599$   & $0.080$ & $0.248$ & $0.458$ & $0.569$   & $0.033$ & $0.064$ & $0.195$ & $0.275$  \\
{\bf KGCN} & $0.190$ & $0.248$ & $0.470$ & $0.603$  & $0.110$ & $0.242$ & $0.462$ & $0.588$   & $0.032$ & $0.071$ & $0.192$ & $0.271$ \\
{\bf VGAE} & $0.190$ & $0.252$ & $0.478$ & $0.597$  & $0.112$ & $0.262$ & $0.473$ & $0.585$   & $0.033$ & $0.081$ & $0.202$ & $0.280$ \\
%{\bf DisenGCN} &  &  &  &    &  &  &  &    &  &  &  &   \\
{\bf Splitter} & $0.185$ & $0.242$ & $0.472$ & $0.589$  & $0.093$ & $0.238$ & $0.462$ & $0.550$   & $0.033$ & $0.061$ & $0.189$ & $0.271$ \\
\hline
{\bf Proposed} & $0.210$ & $0.248$ & $0.500$ & $0.620$  & $0.128$ & $0.281$ & $0.514$ & $0.615$   & $0.050$ & $0.089$ & $0.200$ & $0.275$  \\
\bottomrule
\end{tabular}
\vspace{4pt}
\caption{Recommendation performance comparison in $Recall@k$.} \label{table:recallk}
\vspace{-8pt}
\end{table*}

\subsection{Baseline Methods}
We compare the proposed model with the most representative state-of-the-art methods described below.
\begin{itemize}[leftmargin=*]
    \item \textbf{NMF}~\cite{Koren-etal09MF} is a benchmark collaborative filtering model for recommendation, based on non-negative matrix factorization. The number of factors is chosen as $16$, $16$, $12$ for MovieLens, LastFM and Yelp. $L_1$ and $L_2$ regularizations are weighted equally, with regularization coefficient of $0.005$.
    \item \textbf{FM}~\cite{rendle2010factorization} is a benchmark factorization model which combines second-order feature interactions with linear modeling. We include user IDs, item IDs and items' associated knowledge entities as input. The total dimension of embedding is the same as the proposed model.
    \item \textbf{CKE}~\cite{zhang2016collaborative} extends collaborative filtering with side knowledge information as regularization for recommendations. We use both user-item interactions and knowledge entities as input. The weight for knowledge entity part is set as 0.2 for all datasets. 
    %The total dimension of embedding is the same as the proposed model.
    \item \textbf{RippleNet}~\cite{Wang-etal18ripplenet} is a memory-network-like approach which propagates from items to other entities in the knowledge graph. We set $dim=16$, $H=2$, $\lambda_1=10^{-6}$, $\lambda_2=0.01$, $\eta=0.01$ for MovieLens; $dim=16$, $H=3$, $\lambda_1=10^{-5}$, $\lambda_2=0.02$, $\eta=0.005$ for LastFM; $dim=16$, $H=2$, $\lambda_1=10^{-7}$, $\lambda_2=0.02$, $\eta=0.01$ for Yelp.
    \item \textbf{KGCN}~\cite{Wang-etal19KGCN} is a GCN-based model that captures inter-item relations by mining their attributes on a knowledge graph (KG). We set $dim=120$, $H=2$, $\lambda=10^{-7}$, $\eta=2\times10^{-2}$ for MovieLens; $dim=64$, $H=1$, $\lambda=2\times 10^{-5}$, $\eta=2\times 10^{-4}$ for LastFM; $dim=64$, $H=2$, $\lambda=5\times 10^{-7}$, $\eta=5\times 10^{-4}$ for Yelp.
    \item \textbf{VGAE}~\cite{Kipf-Welling16VGAE} is built upon a graph variational autoencoder. It can be regarded as the fundamental version of the proposed model where the latent dimensions are not disentangled. The hyperparameter setting is similar to the proposed method, except $dim=120$ for MovieLens, $dim=64$ for LastFM, and $dim=64$ for Yelp. The learning rate is $5\times 10^{-4}$ for MovieLens, $2\times 10^{-4}$ for LastFM, and $3\times 10^{-4}$ for Yelp.
    \item \textbf{Splitter}~\cite{Epasto-etal19isaSingle} learns multiple embedding vectors for each node. A clustering step is performed independently as factor identification before embedding learning. Another similar work is~\cite{Liu-etal19isaSingle}. We use overlapping community detection for user-item subgraph and item-entity subgraph. The community detection results are used for factor assignment. Then, we apply $C$ GCN models to learn several embeddings for each node. The community assignments remain fixed. The resultant embeddings are then concatenated for joint prediction. The embedding dimensions and the number of communities are set the same as the proposed model. 
    %$D=30, C=4$ for MovieLens, $D=16, C=4$ for LastFM, and $D=25, C=6$ for Yelp.
\end{itemize}

\begin{table}[t!]
\centering
\vspace{0pt}
\begin{tabular}{@{}c|ccc@{}}\toprule
\multicolumn{1}{c}{} &
\multicolumn{1}{c}{\textbf{MovieLens}} & \multicolumn{1}{c}{\textbf{\,\,\,LastFM\,\,\,}} & \multicolumn{1}{c}{\textbf{Yelp}}\\ 
\midrule
{\bf NMF} & $0.2920$ & $0.1983$ & $0.0681$ \\
{\bf FM} & $0.3020$ & $0.2383$ & $0.0820$ \\
{\bf CKE} & $0.3064$ & $0.2392$ & $0.0959$ \\
{\bf RippleNet} & $0.3153$ & $0.2321$ & $0.1078$ \\
{\bf KGCN} & $0.3230$ & $0.2493$ & $0.1168$ \\
{\bf VGAE} & $0.3234$ & $0.2535$ & $0.1180$ \\
{\bf Splitter} & $0.3242$ & $0.2331$ & $0.1039$ \\
\hline
{\bf Proposed} & $0.3427$ & $0.2656$ & $0.1202$ \\
\bottomrule
\end{tabular}
\vspace{4pt}
\captionof{table}{Recommendation evaluation in $NDCG@100$.} \label{table:ndcg}
\vspace{-10pt}
\end{table}

\subsection{Recommendation Performance Evaluation}
We first compare the performance of the proposed method with the baseline methods. The performances are presented in Table~\ref{table:recallk} and Table~\ref{table:ndcg}. Some observations are drawn below:
\begin{itemize}[leftmargin=*]
    \item The proposed model is at least comparable to the best performance in most cases. VGAE can be seen as the variant of the proposed model without resolving representation learning. It thus shows that considering interpretability will not negatively affect recommendation performances.
    \item Although also emphasizing the idea of disentangled representation learning, the proposed model achieves better performance than Splitter. It demonstrates the advantage of simultaneously learning embedding and identifying factors over treating them as independent steps. The embedding learning in Splitter could be negatively affected by the gap of factor activation between training and inference.
    \item VGAE achieves slightly better performance than KGCN. As we ignore link types in graphs, which is not our focus in this work, the improvement could be from the variational autoencoder.
    \item NMF does not utilize attribute entities, which has a negative impact on its performance, but it serves as the baseline to reflect the degree that attribute information helps in recommendation.
\end{itemize}
% \textcolor{red}{DisenGCN has been shown to improve inductive learning, the scenario of which is different from recommendation.} 

\begin{table}[t!]
\centering
\vspace{0pt}
\begin{tabular}{@{}c|ccc@{}}\toprule
\multicolumn{1}{c}{\textbf{$C\,\,\,$}} &
\multicolumn{1}{c}{\textbf{MovieLens}} & \multicolumn{1}{c}{\textbf{\,\,\,\,\,\,\,\,LastFM\,\,\,}} & \multicolumn{1}{c}{\textbf{Yelp}}\\ 
\midrule
{\bf $c_0-2$} & $0.3398/0.3437$ & $0.2541/0.2722$ & $0.1170/0.1267$ \\
{\bf $c_0-1$} & $0.3422/0.3439$ & $0.2613/0.2754$ & $0.1195/0.1234$ \\
{\bf $c_0$} & $0.3427/0.3427$ & $0.2656/0.2656$ & $0.1202/0.1202$ \\
{\bf $c_0+1$} & $0.3429/0.3308$ & $0.2721/0.2547$ & $0.1204/0.1098$ \\
\bottomrule
\end{tabular}
\vspace{4pt}
\captionof{table}{Performance of the proposed model in $NDCG@100$ by varying the number of factors $C$, where $c_0 = 4, 4, 6$ for MovieLens, LastFM and Yelp, respectively.} \label{table:model_k} 
\vspace{-8pt}
\end{table}

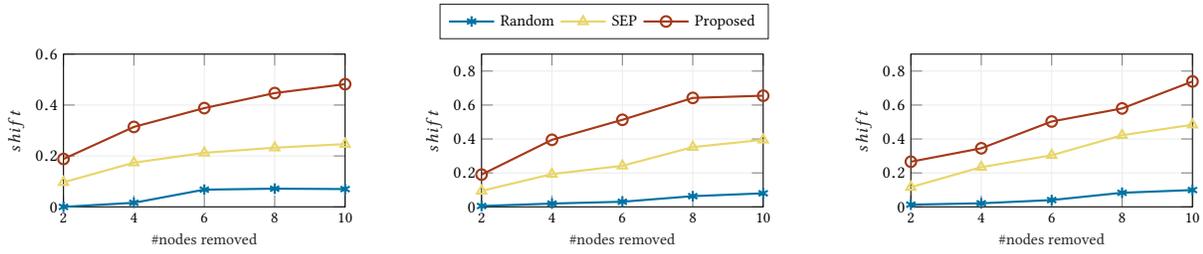
\begin{figure*}[t]
\captionsetup[subfigure]{justification=centering}
\hspace{-25pt}
\begin{subfigure}[b]{0.3\textwidth}
  \setlength\figureheight{0.80in}
  \setlength\figurewidth{1.55in}
  \centering  \scriptsize
      \input{figures/MovieLens_adv.tikz} 
  %\caption{Level 1}
  %\label{fig:NGNMF}
\end{subfigure}%
\hspace{4pt}
\begin{subfigure}[b]{0.3\textwidth}
  \setlength\figureheight{0.80in}
  \setlength\figurewidth{1.55in}
  \centering  \scriptsize
      \input{figures/LastFM_adv.tikz} 
  %\caption{Level 2}
  %\label{fig:NGNMF}
\end{subfigure}
\hspace{6pt}
\begin{subfigure}[b]{0.3\textwidth}
  \setlength\figureheight{0.80in}
  \setlength\figurewidth{1.55in}
  \centering  \scriptsize
      \input{figures/Dianping_adv.tikz} 
  %\caption{Level 2}
  %\label{fig:NGNMF}
\end{subfigure}
\vspace{-5pt}
\caption{Quantitative evaluation of explanation faithfulness on MovieLens (left), LastFM (middle) and Yelp (right).}
\vspace{-0pt}
\label{fig:adv_exp}
\end{figure*}

\subsubsection{Hyperparameter Sensitivity Analysis}
The most important hyperparameter in this work is the number of factors $C_1$ and $C_2$. Therefore, we further analyze model performances by changing factor numbers. Here we let $C_1=C_2=C$. Results are reported in Table~\ref{table:model_k}. For each pair of values, the left value means that we fix the dimension $D$ of each embedding segment, while the right value means that we fix the total embedding dimension $C\times D$.

We could observe that, by increasing $C$, model performance in general increases if we fix $D$, since more factors are considered and the parameter size also increases. The time complexity and memory cost will also increase in this case. However, the model performance decreases if we set $C$ too large while keeping $C\times D$ fixed, since the dimension for each factor decreases and is not adequate to contain all the information needed.

% Note: The model involved in this part does not have self-embedding connection in the bottom layer. Please check xxx_adv.py of the proposed implementation.
\subsection{Evaluation of Explanation Accuracy}\label{subsec:eval_intp}
In this part, we will first evaluate the faithfulness of interpretation obtained from our model. Then, we qualitatively analyze the insights from interpretation. Explanation information comes from two parts: (1) affiliation scores tracing the information flows; (2) factors extracted on different embedding segments.

\subsubsection{Quantitative Evaluation of Explanation} Evaluating the correctness of model explanation quantitatively is a challenging task, because usually there is no ground-truth explanation to compare with. To tackle this, we utilize the duality between interpretation and adversarial perturbation~\cite{Fong-Vedaldi17perturbation}. The high-level intuition is that, after removing the features that are important to the current prediction, the model prediction should dramatically change.

Specifically, we use the $200$ users in testing data for evaluation. After performing recommendation, interpretation is obtained as the historical items or important attribute entities. For each user-item pair $(u,t)$, we compute the importance score of an item/entity $j$ as $s(j;u, t)$. The exact definition of $s(j;u, t)$ is introduced in Appendix. In brief, $s(j;u, t)$ is computed based on the affiliation scores $p(e,c)$ and $p(t,c)$ as they trace contribution from entities to items and from items to users. Then, the items/entities with large importance scores are removed from the original input. Finally, we perform recommendations again with the new input. Let $recall$ and $recall'$ denote the $Recall@10$ performance before and after the perturbation, respectively. We define $shift = \frac{recall - recall'}{recall}$ to measure the explanation accuracy. A higher value indicates more accurate explanation.

We also introduce two baseline methods to compare with the proposed interpretation scheme. Popular methods such as LIME~\cite{Ribeiro-etal16whyshould} is not included since there is no explicit concept of classes in recommendation settings.
\begin{itemize}[leftmargin=*]
    \item \textbf{Random}: We randomly remove the same number of historical items as the proposed method.
    \item \textbf{SEP}~\cite{yang2018towards}: A post-hoc path-based interpretation method for recommender systems. Items in the resultant explanation paths are retrieved as interpretation.
\end{itemize}

The experimental results are shown in Figure~\ref{fig:adv_exp}. We make $5$ runs for each case and take the average. We vary the number of items to remove in $x$-axis, and the $shift$ is $y$-axis. Observations are summarized as follows:
\begin{itemize}[leftmargin=*]
    \item The interpretation obtained directly from model inference is more accurate than using post-hoc explanations. It thus validates the benefit of designing explainable models over using external explanation methods.
    \item The amount of information removed is small compared with the overall input. Also, the $shift$ value increases as the amount of removed information increases. It thus further demonstrates the faithfulness of the obtained explanations.
\end{itemize}

\begin{figure}[t]
\centering
\includegraphics[scale=0.34]{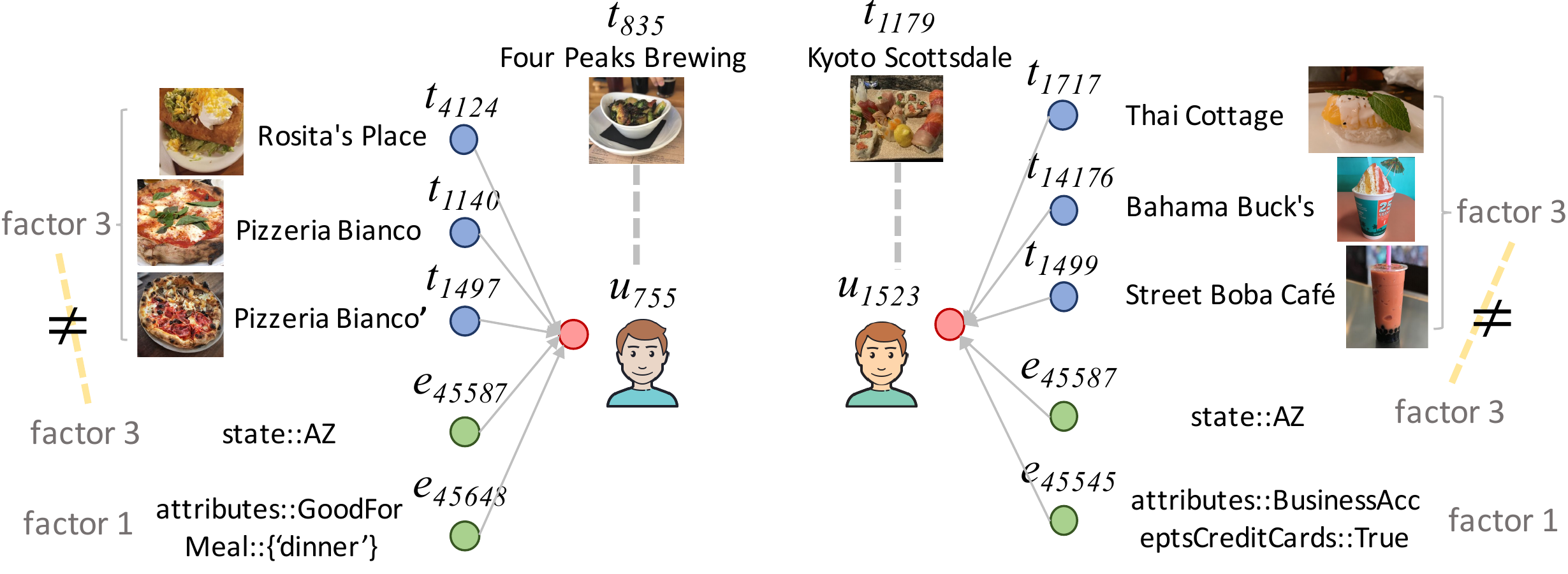}
\vspace{-10pt}
\caption{Visualization of the explanations.}
\label{fig:case2}
\vspace{-6pt}
\end{figure}

\subsubsection{Case Studies}
After the quantitative evaluation, we provide some visualization of explanation results in Figure~\ref{fig:case2}. We can see that restaurant $t_{835}$ is recommended to user $u_{755}$ because the user has visited similar restaurants (near Phoenix) and his activities are mainly in Arizona emphasized by attribute information. Similarly, restaurant $t_{1179}$ is recommended to user $u_{1523}$ because he has visited many Asian restaurants. The factor that each node is affiliated with is also shown.

We also observe an interesting phenomenon in the second recommendation. The recommendation is partially explained by the attribute $e_{45545}$, which indicates that the restaurant accepts credit cards for payment. However, intuitively it may not be a good reason shown to users to explain recommendation results in practice. The reason that this attribute is selected could be that many restaurants accept credit cards, but the strong correlation is not suitable as the reason to persuade users. This could be a limitation for graph convolution models. It also reminds us that interpretability is not equal to credibility, and we leave it for future exploration.

\subsubsection{Visualization of Latent Dimensions}
\begin{figure}[t]
\centering
\includegraphics[scale=0.16]{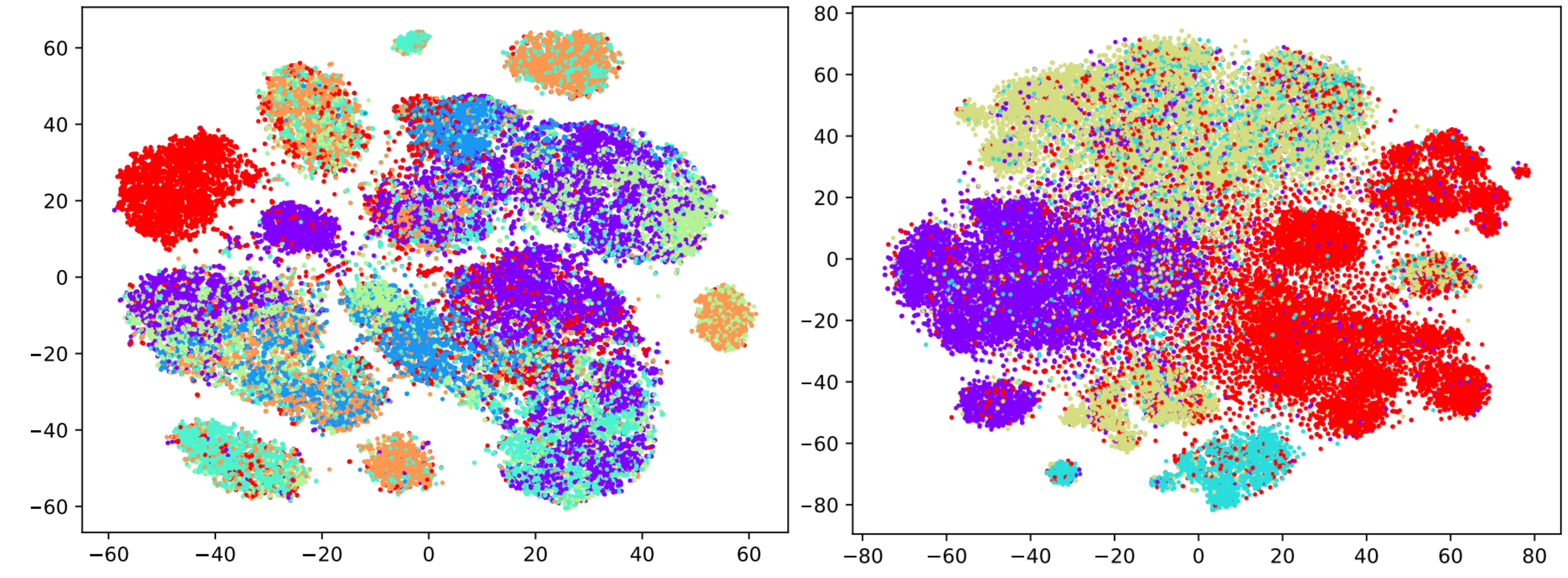}
\vspace{-4pt}
\caption{Visualization of embeddings for Yelp dataset.}
\label{fig:embd_vis}
\vspace{-10pt}
\end{figure}

In this part, we visualize the item embeddings in Yelp dataset in Figure~\ref{fig:embd_vis}. The left plot follows the previous settings where $C_1=6$, while the right plot is drawn as we let $C_1=4$. The color of each point indicates its closest affiliated factor. Some colors are mixed since an embedding could be assigned to multiple factors, but only one factor could be shown for each embedding. The clustering phenomenon is consistent with factor affiliations. It demonstrates that our model is able to discover different factors from data and assign embeddings accordingly.

% Analyze the frequent nodes in each set of dimension.

% Analyze the correlation between different dimensions~\cite{Ma-etal19disentGCN}.

% For those links learned by the model, what is the distribution of similarity value of each dimension between a pair of user and item?

%% file: figures/MovieLens_adv.tikz
\definecolor{mycolor1}{rgb}{0.00000,0.44700,0.64100}%
\definecolor{mycolor3}{rgb}{0.65000,0.22500,0.09800}%
\definecolor{mycolor4}{rgb}{0.91, 0.84, 0.42}
\definecolor{mycolor2}{rgb}{0.14700,0.59800,0.17500}%

\begin{tikzpicture}

\begin{axis}[%
width=0.951\figurewidth,
height=\figureheight,
at={(0\figurewidth,0\figureheight)},
scale only axis,
scaled x ticks=true,
xticklabels={2,4,6,8,10},
xtick={1,2,3,4,5},
xmin=1,
xmax=5,
xlabel style={font=\color{white!15!black}},
xlabel={\#nodes removed},
ymin=0.00,
ymax=0.60,
grid, % --added
grid style={line width=.15pt, draw=gray!15}, % --added, dashed, 
ylabel style={font=\color{white!15!black}},
ylabel={$shift$},
axis background/.style={fill=white},
axis background/.style={fill=white} % --added
]
\addplot [color=mycolor1, mark=asterisk, mark options={solid, mycolor1}, thick]
  table[row sep=crcr]{%
1	0.0\\
2	0.01595\\
3	0.06759\\
4	0.07143\\
5	0.06986\\
};
%\addlegendentry{RAND}

\addplot [color=mycolor4, mark=triangle, mark options={solid, mycolor4}, thick]
  table[row sep=crcr]{%
1	0.09652\\
2	0.17354\\
3	0.2122\\
4	0.23245\\
5	0.24652\\
};
%\addlegendentry{SEP}

\addplot [color=mycolor3, mark=o, mark options={solid, mycolor3}, thick]
  table[row sep=crcr]{%
1	0.188\\
2	0.314\\
3	0.388\\
4	0.447\\
5	0.482\\
};
%\addlegendentry{INTP}

\end{axis}
\end{tikzpicture}%

%% file: figures/LastFM_adv.tikz
\definecolor{mycolor1}{rgb}{0.00000,0.44700,0.64100}%
\definecolor{mycolor3}{rgb}{0.65000,0.22500,0.09800}%
\definecolor{mycolor4}{rgb}{0.91, 0.84, 0.42}
\definecolor{mycolor2}{rgb}{0.14700,0.59800,0.17500}%

\begin{tikzpicture}

\begin{axis}[%
width=0.951\figurewidth,
height=\figureheight,
at={(0\figurewidth,0\figureheight)},
scale only axis,
scaled x ticks=true,
xticklabels={2,4,6,8,10},
xtick={1,2,3,4,5},
xmin=1,
xmax=5,
xlabel style={font=\color{white!15!black}},
xlabel={\#nodes removed},
ymin=0.00,
ymax=0.90,
grid, % --added
grid style={line width=.15pt, draw=gray!15}, % --added, dashed, 
ylabel style={font=\color{white!15!black}},
ylabel={$shift$},
axis background/.style={fill=white},
legend columns = 3,
legend style={legend cell align=left, align=left, draw=white!15!black, nodes={scale=1.0}, at={(1.02, 1.33)}},
axis background/.style={fill=white} % --added
]
\addplot [color=mycolor1, mark=asterisk, mark options={solid, mycolor1}, thick]
  table[row sep=crcr]{%
1	0.00495\\
2	0.01935\\
3	0.030175\\
4	0.063\\
5	0.07975\\
};
\addlegendentry{Random}

\addplot [color=mycolor4, mark=triangle, mark options={solid, mycolor4}, thick]
  table[row sep=crcr]{%
1	0.094\\
2	0.193\\
3	0.242\\
4	0.352\\
5	0.395\\
};
\addlegendentry{SEP}

\addplot [color=mycolor3, mark=o, mark options={solid, mycolor3}, thick]
  table[row sep=crcr]{%
1	0.190\\
2	0.395\\
3	0.513\\
4	0.642\\
5	0.655\\
};
\addlegendentry{Proposed}

\end{axis}
\end{tikzpicture}%

%% file: figures/Dianping_adv.tikz
\definecolor{mycolor1}{rgb}{0.00000,0.44700,0.64100}%
\definecolor{mycolor3}{rgb}{0.65000,0.22500,0.09800}%
\definecolor{mycolor4}{rgb}{0.91, 0.84, 0.42}
\definecolor{mycolor2}{rgb}{0.14700,0.59800,0.17500}%

\begin{tikzpicture}

\begin{axis}[%
width=0.951\figurewidth,
height=\figureheight,
at={(0\figurewidth,0\figureheight)},
scale only axis,
scaled x ticks=true,
xticklabels={2,4,6,8,10},
xtick={1,2,3,4,5},
xmin=1,
xmax=5,
xlabel style={font=\color{white!15!black}},
xlabel={\#nodes removed},
ymin=0.00,
ymax=0.90,
grid, % --added
grid style={line width=.15pt, draw=gray!15}, % --added, dashed, 
ylabel style={font=\color{white!15!black}},
ylabel={$shift$},
axis background/.style={fill=white},
axis background/.style={fill=white} % --added
]
\addplot [color=mycolor1, mark=asterisk, mark options={solid, mycolor1}, thick]
  table[row sep=crcr]{%
1	0.013\\
2	0.021\\
3	0.040\\
4	0.083\\
5	0.099\\
};
%\addlegendentry{RAND}

\addplot [color=mycolor4, mark=triangle, mark options={solid, mycolor4}, thick]
  table[row sep=crcr]{%
1	0.116\\
2	0.234\\
3	0.304\\
4	0.422\\
5	0.484\\
};
%\addlegendentry{SEP}

\addplot [color=mycolor3, mark=o, mark options={solid, mycolor3}, thick]
  table[row sep=crcr]{%
1	0.266\\
2	0.345\\
3	0.503\\
4	0.580\\
5	0.739\\
};
%\addlegendentry{DFE}

\end{axis}
\end{tikzpicture}%

%% file: 6_RelatedWork.tex
\textbf{Machine Learning for Recommender Systems.} In addition to traditional content-based and neighborhood-based methods~\cite{GeIJCAI20162,GeTOIS20141}, advanced machine learning methods have been developed for recommendation, which could be categorized into two groups: linear and nonlinear. Examples of linear methods include matrix factorization (MF) and factorization machines (FMs). Specifically, MF~\cite{baltrunas2011matrix} represents an individual user or item by a single latent vector and models the interaction of a user-item pair by the inner product of latent vectors. FMs~\cite{rendle2010factorization} embed features into a latent space and model user-item interactions by summing up the inner products of embedding vectors between all pairs of features. To this end, a lot of recent efforts have been put on modeling user-item interactions in a nonlinear way such as using DNNs. Example studies include the Wide \& Deep model for app recommendation in Google Play~\cite{Cheng-etal16wideanddeep}, the non-linear extensions of MF and FM~\cite{he2017factorization,He-etal17NCF}, the recommender modeled by recurrent neural networks (RNNs)~\cite{donkers2017sequential}, etc. It has been demonstrated that these nonlinear models usually yield better effectiveness, where a comprehensive review is available in~\cite{zhang2019deep}.  

\noindent \textbf{Interpretable Machine Learning.} Model interpretability receives increasing attention recently~\cite{Montavon-etal18survey, Du-etal18techniques}. The efforts can be divided into two categories, post-hoc interpretation methods and intrinsically interpretable models. Specifically, post-hoc methods can be further divided into global and local methods, where the former transforms black-box models into understandable structures~\cite{ZhangQ-etal18cnnTrees}, and the latter explains individual predictions~\cite{Ribeiro-etal16whyshould, Simonyan-etal13deepInsideCNNsaliency}. There are gradient-based methods that measure feature contributions on output or intermediate neurons through backpropagation~\cite{Simonyan-etal13deepInsideCNNsaliency}, perturbation-based methods~\cite{Fong-Vedaldi17perturbation}, approximation-based methods~\cite{Ribeiro-etal16whyshould}, and entropy-based methods~\cite{Guan-etal19unifiedNLP}. Intrinsically interpretability focuses on developing transparent model components. A typical strategy is to pose regularization on parameters to conform with certain prior~\cite{Higgins-etal17betaVAE}. \cite{Wang-etal18credibleModel} proposes to incorporate expert knowledge to penalize model weights. \cite{Tsang-etalNIPS18interactionTransparency} develops disentangled MLPs to explicitly regulate feature interactions between layers.

\noindent \textbf{Explainable Recommender Systems}. The explainability of recommender systems recently starts to be seen as an important topic~\cite{Zhang-Chen17explainableRec}. Before the popularity of deep models, explicit factor models have been proposed to relate explicitly constructed features with latent factors as explanations for the factors~\cite{Zhang-etal14explictFactor}. To balance effectiveness and explainability, a common strategy is to feed interpretable models with engineered features~\cite{he2014practical} or construct domain-specific knowledge graphs~\cite{Wang-etal19AAAIrecomm}. Graphs can be used as the basis to pass information for embedding learning~\cite{Wang-etal18ripplenet, Wang-etal19reasoningKG, Wang-etal19KGAT}. \cite{hu2018interpretable} proposes an attention based model, but it only works on text data. \cite{ma2019learning} proposes a new recommendation model to learn disentangled representations, but how to apply it into general scenarios containing extra attribute information and knowledge is beyond the discussion. To discriminate different semantics in embedding, \cite{Liu-etal19isaSingle, Epasto-etal19isaSingle} extend random-walk-based models and use multiple embeddings to represent each node, but embedding learning and factor extraction are done separately.
% Recently developed models usually map items, features and users into an embedding space for similarity measurement~\cite{Ying-etal18GCNrecomm, Zhou-etal18attentionCTR, Wang-etal19KGCN, Wang-etal19KGAT}.

%% file: 7_Conclusion.tex
In this work, we tackle the explainability problem in recommender systems. Specifically, we first analyze the IOM elements that benefit model interpretability, and then propose a novel model with a more transparent representation learning module. The IOM elements include reshaping input into discrete concepts, making output attributable, and disentangling the middle-level representations. The proposed model considers all the three elements above. The data is formatted as a graph. The model resolves information passing in graph convolution according to different source types and factors. Different from previous work, in our model, the representation learning process and factor identification are achieved simultaneously. Experiments on real-world datasets validate the effectiveness and interpretation accuracy of our model.

The future work includes: (1) developing more effective graph construction methods for automatic input conceptualization; (2) designing human-computer interaction interfaces to better render interpretation results to users; (3) extending the proposed model to handle a larger number of facets.